\documentclass{article}
\usepackage{spconf,amsmath,graphicx}
\usepackage{siunitx}
\usepackage{multirow}
\usepackage{graphicx}
\usepackage{hyperref}

\title{Automatic Error Detection in Integrated Circuits Image Segmentation: A Data-driven Approach}
\name{\textit{Zhikang Zhang} $^{1}$ \qquad \textit{Bruno Machado Trindade} $^{2}$ \qquad \textit{Michael Green} $^{2}$ \qquad \textit{Zifan Yu} $^{1}$ \\ \textit{Christopher Pawlowicz} $^{2}$ \qquad \textit{Fengbo Ren}$^{1}$\thanks{This work is supported by a research contract from TechInsights Inc. Corresponding author: Zhikang.Zhang@asu.edu} }
\address{$^{1}$ Arizona State University, Tempe, AZ, USA \qquad $^{2}$ TechInsights Inc., Ottawa, ON, Canada}

\begin{document}
%
\maketitle
\begin{abstract}
 Due to the complicated nanoscale structures of current integrated circuits(IC) builds and low error tolerance of IC image segmentation tasks, most existing automated IC image segmentation approaches require human experts for visual inspection to ensure correctness, which is one of the major bottlenecks in large-scale industrial applications. In this paper, we present the first data-driven automatic error detection approach targeting two types of IC segmentation errors: wire errors and via errors. On an IC image dataset collected from real industry, we demonstrate that, by adapting existing CNN-based approaches of image classification and image translation with additional pre-processing and post-processing techniques, we are able to achieve recall/precision of 0.92/0.93 in wire error detection and 0.96/0.90 in via error detection, respectively. 
\end{abstract}
\begin{keywords}
error detection, reverse engineering, image segmentation, image classification, image translation
\end{keywords}
\section{Introduction}
\begin{figure}
    \centering
    \includegraphics[width=.75\linewidth]{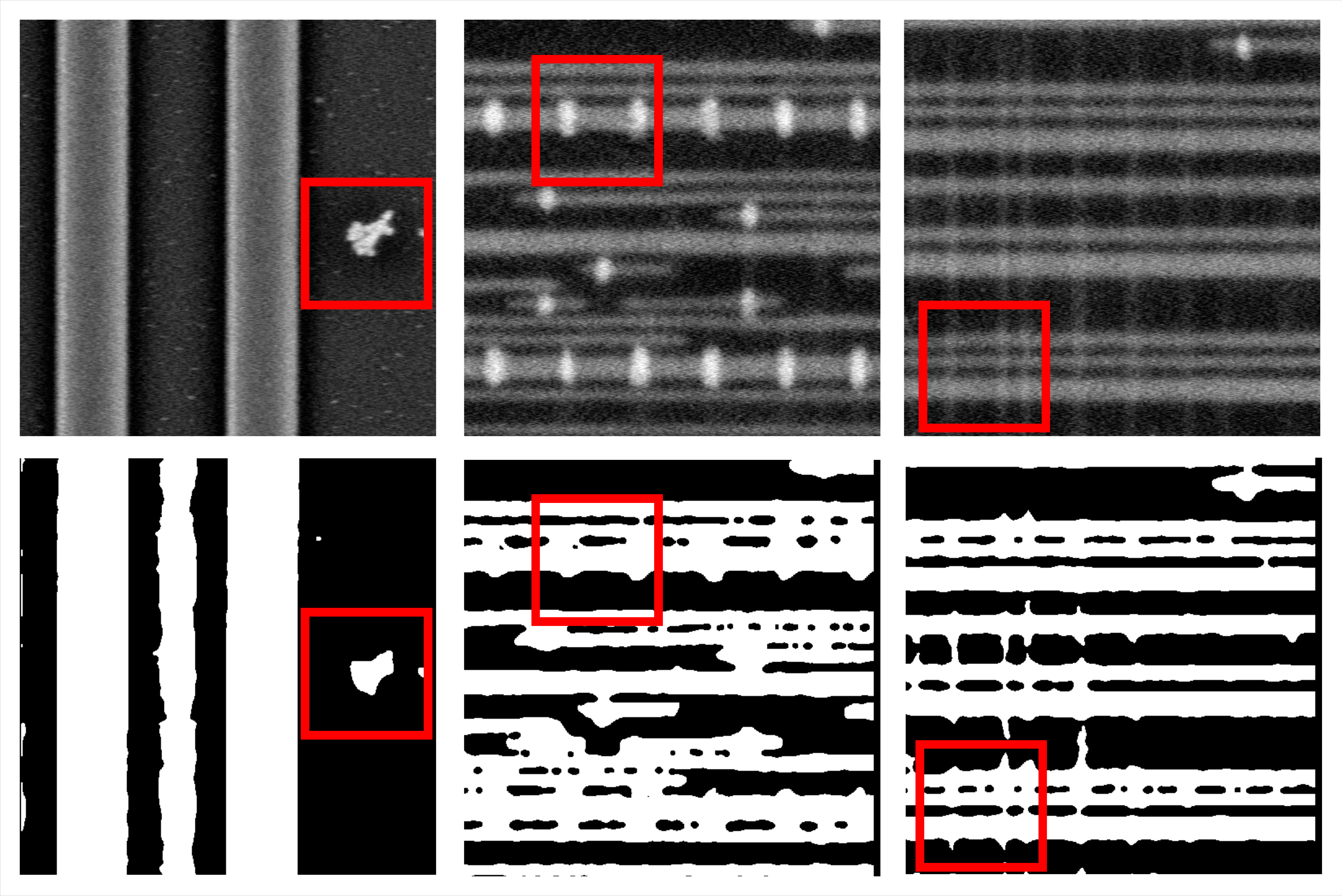}
    \caption{Image samples with errors caused by unexpected factors. Row 1: SEM images. Row 2: Wire segmentation results. Column 1: Random Contamination. Column 2: Improper exposure. Column 3: Improper layer removal.}
    \label{errorlook}
\end{figure}

The ever-developing integrated circuits(IC) manufacturing technologies lead to the ever-increasing complexity of IC structures built at the nanoscale. Such high complexity in IC can put normal users and IC designers at risk by providing opportunities for adversaries to hide malicious or IP-protected designs within IC\cite{Wilson2022REFICSAS}. Reverse engineering is the best or even the only approach up to date that can help address this problem\cite{Wilson2022REFICSAS}. In a nutshell, reverse engineering operates by repeatedly imaging and physically removing the topmost layer of an IC chip to reveal all the inner 3D structures for analysis. Due to the extremely small size of IC structures, scanning electron microscopy(SEM) is often used as an imaging tool. Once SEM images are acquired, the coarse IC designs can be retrieved by segmenting SEM images into different objects(we only target vias and wires in this work). The SEM image quality can be affected by various unexpected factors, such as random contaminations, improper exposures, improper layer removal, etc, which lead to unexpected segmentation errors for existing SEM image segmentation approaches\cite{Hong2018DeepLF,Cheng2018Hybrid,Yu2022ADA,Cheng2019AHM,Cheng2019GlobalTP,Wilson2020LASREAN,Doudkin2005ObjectsIO,Wilson2020HistogrambasedAS,Lee2008AnEI,Lagunovsky1998RecognitionOI}, as shown in Fig.~\ref{errorlook}.). To correct such errors, human experts are required to visually inspect the segmentation results. However, in real world, up to thousands of SEM images are generated from a single IC chip\cite{Wilson2022REFICSAS}, which makes manual inspection highly challenging and laborious, leading to one of the major bottlenecks in large-scale industrial applications.

In this work, we target the problem of automatic via and wire error detection in segmented SEM images. We formulate the wire error detection and via error detection as image classification and image translation problems, respectively. By adapting existing related approaches with additional image pre-processing and post-processing techniques, we achieve high performances in both via and wire error detection problems on an SEM image dataset we collect from the real industry. 
\begin{figure*}
    \centering
    \includegraphics[width=.9\textwidth]{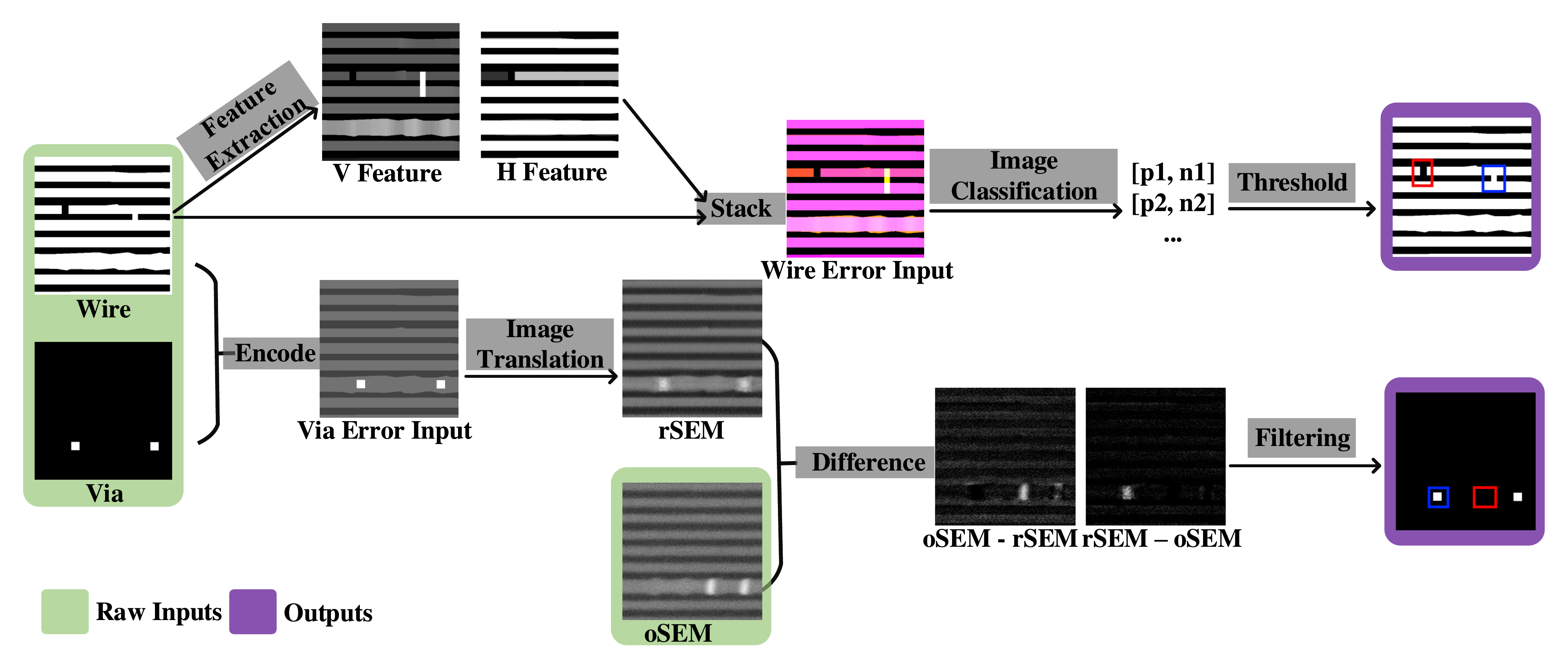}
    \caption{The pipeline of our approach. rSEM: reconstructed SEM image. oSEM: original SEM image. Red bounding boxes: "open" of wires or "miss" vias. Blue bounding boxes: "short" of wires or "extra" vias}
    \label{pipeline}
\end{figure*}

We summarize our contributions as follows:

1. We propose the first data-driven approach targeting error detection in segmented SEM images of IC. Our approach achieves high performance on both via and wire error detection, which significantly unblocks a major bottleneck for SEM image segmentation approaches.

2. We demonstrate that existing work in image classification and image translation can be well adapted for error detection in SEM image segmentation to achieve high performances with the help of necessary image pre-processing and post-processing techniques.

3. We evaluate our approach on a real industrial dataset and achieve an average recall/precision of 0.92/0.93 in wire error detection and 0.96/0.90 in via error detection, respectively. The high performance in error detection implies the potentially broad applicability of our approach.

\section{Methodology}
\subsection{Datasets and Evaluation Metrics}
\label{section31}
\textbf{Datasets.} We collect 39 SEM images of various hardware: microprocessor, radio frequency(RF) transceiver, power management, flash memory, SoC, and so on, with a collecting dwelling time of 0.2 $\mu$s/pixel. The average pixel size and field sizes are 2.92$nm$ and 22.96 $\mu$m, respectively. For brevity, this SEM image set is denoted as S0. Each image in S0 is in grayscale with a size of 8192x8192, which is too high to feed into common CNNs. Thus we choose to process images at patch level (256x256) first and then merge patch-level results to form full image results. There are two types of objects to segment: vias and wires. Vias are electrical connections between copper layers in ICs, which are often imaged with the highest pixel intensity and shown as small rounded squares. Wires are imaged with lower pixel intensity than vias but higher pixel intensity than backgrounds and shown as long strips(Fig.~\ref{pipeline}.). We collect four wire segmentation sets(denoted as W0, W1, W2, and W3) and two via segmentation sets(denoted as V0 and V1) corresponding to S0. W0 set is generated by \cite{Trindade2018SegmentationOI} with manual corrections. W1, W2, and W3 sets are generated by \cite{Yu2022ADA} using three different settings. V0 and V1 sets are generated by \cite{Trindade2018SegmentationOI} and \cite{Yu2022ADA}, respectively.

\textbf{Evaluation metrics.} In the context of SEM image segmentation of IC, if and only if the segmentation differences that cause errors in connectivity of IC are defined as segmentation errors. More precisely, for wires, only an "open" or a "short" are errors(Fig.~\ref{pipeline}.). For vias, only a "miss" or an "extra" are errors(Fig.~\ref{pipeline}.). To evaluate the error detection performance of our approach, we use the following metrics: for wire segmentation errors, given a wire error segmentation result EW and the corresponding ground truth segmentation result GW, we denote each patch of the image as either "correct" or "error" based on whether there is an electric-significant-differences(ESD)\cite{Trindade2018SegmentationOI}in the patch. The wire error detection performance is quantified by the number of detected error patches. For via segmentation errors, given a via error segmentation result EV and the corresponding ground truth segmentation result GV, we first extract all isolated regions from both EV and GV using \cite{wu2005optimizing}. Then each region from EV that overlaps with a region from GV is treated as a correctly segmented via. Other regions from either EV or GV that has no overlapping regions from GV or EV are treated as errors(corresponding to extra vias and miss vias, respectively). The via error detection performance is quantified by the number of detected via errors. 

\subsection{Wire Error Detection}
We formulate the wire error detection problem as an image classification problem. A CNN-based binary image classifier slides over the pre-processed wire segmentation images to determine whether each image patch has errors. 

\begin{figure}
    \centering
    \includegraphics[width=.5\linewidth]{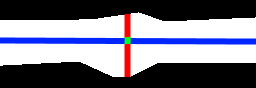}
    \caption{For a given pixel(green), the horizontal and vertical extension values are defined as the length of blue and red lines, respectively.}
    \label{vh}
\end{figure}

\textbf{Pre-processing.} Since the input patch size(256x256) is significantly lower than the original image size(8192x8192), we compose novel features to implicitly encode global information into inputs. More specifically, we first calculate the horizontal and vertical extension values of each pixel(as defined in Fig.~\ref{vh}.) in the full-size wire segmentation image. Then the values are normalized into the $[0,255]$ range to form an H feature and a V feature image, respectively. The original wire segmentation image is stacked with V and H features to form an RGB image as the input of the image classifier, as shown in Fig~.\ref{pipeline}. 

\textbf{Training an image classifier.} 
To compose a training set, we take W0 set as the ground truth and randomly sample 256x256 image patches from W1, W2, and W3 as training samples. Each image patch is labeled as either positive or negative based on whether there is an error. We use ResNet\cite{he2016deep} as the image classifier. The FC layer that generates the final output is modified to generate a binary output. We take the training strategy from \cite{pytorchimagenet} to train the network. Since the training set is highly unbalanced(positive samples are only 3\%), we used a weighted cross-entropy loss whose weights for positive and negative samples are defined as $\frac{N}{P+N}$ and $\frac{P}{P+N}$ where $P$ and $N$ are numbers of positive and negative samples. 

\textbf{Post-processing.}
In the inference stage, for each 256x256 patch, the direct output of the image classifier is a two-element vector denoted as $[p,n]$, where a positive detection is made when $p>n$. We apply a threshold over outputs of multiple overlapping patches to localize errors into smaller areas. The more positive detections are made, the more likely there is an error in the overlapping area. It should be noted that this post-processing step is only needed to localize wire errors more precisely. To detect whether a given 256x256 patch contains errors, using the direct output of the image classifier is sufficient. 

\subsection{Via Error Detection}
We formulate the via error detection as a one-to-one image translation problem. Pix2pix\cite{isola2017image} is used as the image translation method. Given a pair of wire and via segmentation image W and V and corresponding original SEM image(oSEM),  W and V are first encoded into one image and then fed into the pix2pix model to be translated into a reconstructed SEM image(rSEM). At last, the errors are detected by processing the differences between rSEM and oSEM, as shown in Fig.~\ref{pipeline}. To note, W is assumed to contain no errors.

\textbf{Pre-processing.}
We observed that pixel values of via, wire, and background are relatively static within the same SEM image but highly dynamic across different SEM images. Thus we dynamically encode via and wire segmentation images as input to pix2pix model with respect to each SEM image. We first estimate three representative pixel values for via, wire, and background denoted as $v$, $w$, and $b$. $v$, $w$, and $b$ are the 90th percentile, the median, and the 10th percentile of via, wire, and background pixels' values.  The pixels are isolated from oSEM according to via and wire segmentation images V and W. 

\textbf{Training a pix2pix model.}
In our experiments, W0 is used as the wire segmentation set. We compose a training set consisting of 39936 image patches(1024 patches per full-sized image)randomly sampled from both V0 and V1. The training strategy is taken from \cite{pytorchp2p}. Most of the training settings remain consistent with the original work. The differences are as follows: 1. we set the number of input and output channels to one. 2. All the data augmentations are disabled. 3. The total training epochs are set to 10. In the last five epochs, the learning rate decays linearly. 

\textbf{Post-processing.}
We first calculate two differences between oSEM and rSEM: $D1=rSEM-oSEM$ and $D2=oSEM-rSEM$. The negative values in $D1$ and $D2$ are set to zeros. $D1$ and $D2$ are used for detecting extra "vias" and "miss" vias, respectively. We transform positive pixels in $D1$ and $D2$ into isolated regions using \cite{wu2005optimizing}. For each region, we first measure two values: 1. the size of the bounding box covering the region. 2. The average pixel values of nonzero pixels within the region. By setting proper upper and lower thresholds to these values, we can filter out candidate vias. At last, the remaining regions in D1/D2 are marked as extra/miss vias when they overlap/do not overlap with any vias in V.

\section{Experiments}
\textbf{Wire error detection.} We compose two training sets in different sizes, denoted as "small" and "large", respectively. "small" set consists of 229,049 samples with 6883 positive samples. "large" set consists of 1,030,699 training samples with 27833 positive samples. The testing set consists of regularly sampled nonoverlapping image patches from W1, W2, and W3, including 99956 samples with 2437 positive samples. We trained six different models corresponding to different input image encoding, network structures, training set sizes, and network structures. The experiment results are shown in Table~\ref{wireerrorresults}. 
\begin{table}[th]
\centering
\resizebox{\linewidth}{!}{%
\begin{tabular}{|c|c|c|c|c|c|}
\hline
Network  & Input & \begin{tabular}[c]{@{}c@{}}Dataset\end{tabular} & Strategy                                                & Recall & Precision \\ \hline
ResNet18 & W     & small                                               & \begin{tabular}[c]{@{}c@{}}From \\ scratch\end{tabular} & 0.83   & 0.77      \\ \hline
ResNet18 & VH    & small                                               & \begin{tabular}[c]{@{}c@{}}From \\ scratch\end{tabular} & 0.83   & 0.64      \\ \hline
ResNet18 & WVH   & small                                               & \begin{tabular}[c]{@{}c@{}}From \\ scratch\end{tabular} & 0.84   & 0.79      \\ \hline
ResNet34 & WVH   & small                                               & \begin{tabular}[c]{@{}c@{}}From \\ scratch\end{tabular} & 0.82   & 0.76      \\ \hline
ResNet18 & WVH   & large                                               & \begin{tabular}[c]{@{}c@{}}From \\ scratch\end{tabular} & 0.92   & 0.93      \\ \hline
ResNet18 & WVH   & small                                               & Tuning                                                  & 0.61   & 0.30      \\ \hline
\end{tabular}%
}
\caption{The wire error detection performance of various trained models. W: encode wire segmentation only. VH: encode V and H features only. WVH: encode wire segmentation with V and H features together. From scratch: training the network from scratch. Tuning: use a ResNet18 pre-trained on ImageNet and only tune the last FC layer in training.}
\label{wireerrorresults}
\end{table}

The highest error detection performance is a recall/precision of 0.92/0.93 with ResNet-18 trained on the "large" set. The input image is encoded with the wire segmentation result, V and H features. These results indicate that our approach can effectively detect wire segmentation errors without human intervention. 

Compared with other trained models in Table~\ref{wireerrorresults}, we conclude the followings: 1. A deeper network structure does not necessarily lead to higher detection performance(ResNet18 vs. ResNet34, row 3 and row 4). This implies that the wire error detection task relies more on low-level, local information in images, which can be extracted by shallower neural networks. 2. Using wire segmentation images together with V and H features improves the detection performance while using V and H features only is not sufficient(row 1, row 2, and row 3). 3. A larger training set effectively improves detection performance(row 3 and row 5). 4. Pre-training on ImageNet is not necessary(row 3 and row 6), which further implies that wire error detection relies more on low-level information, thus a network trained for a high-level task does not perform well.

\begin{figure}
    \centering
    \includegraphics[width=\linewidth]{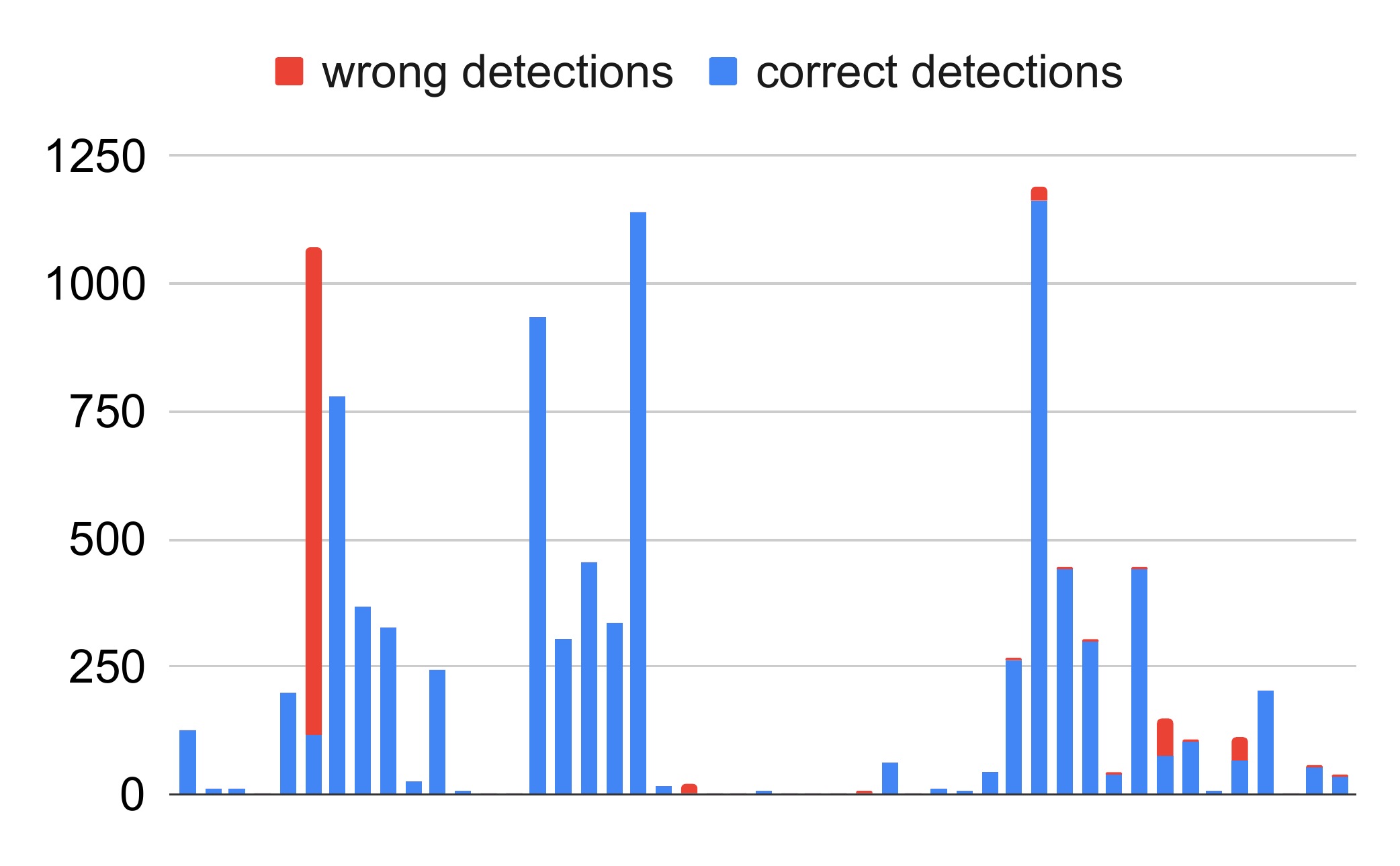}
    \caption{The via error detection results. Wrongly detected errors are either actual errors not detected(false negative results) or no-existing errors detected(false positive results)}
    \label{viaresult}
\end{figure}

\textbf{Via error detection.}
\begin{figure}
    \centering
    \includegraphics[width=\linewidth]{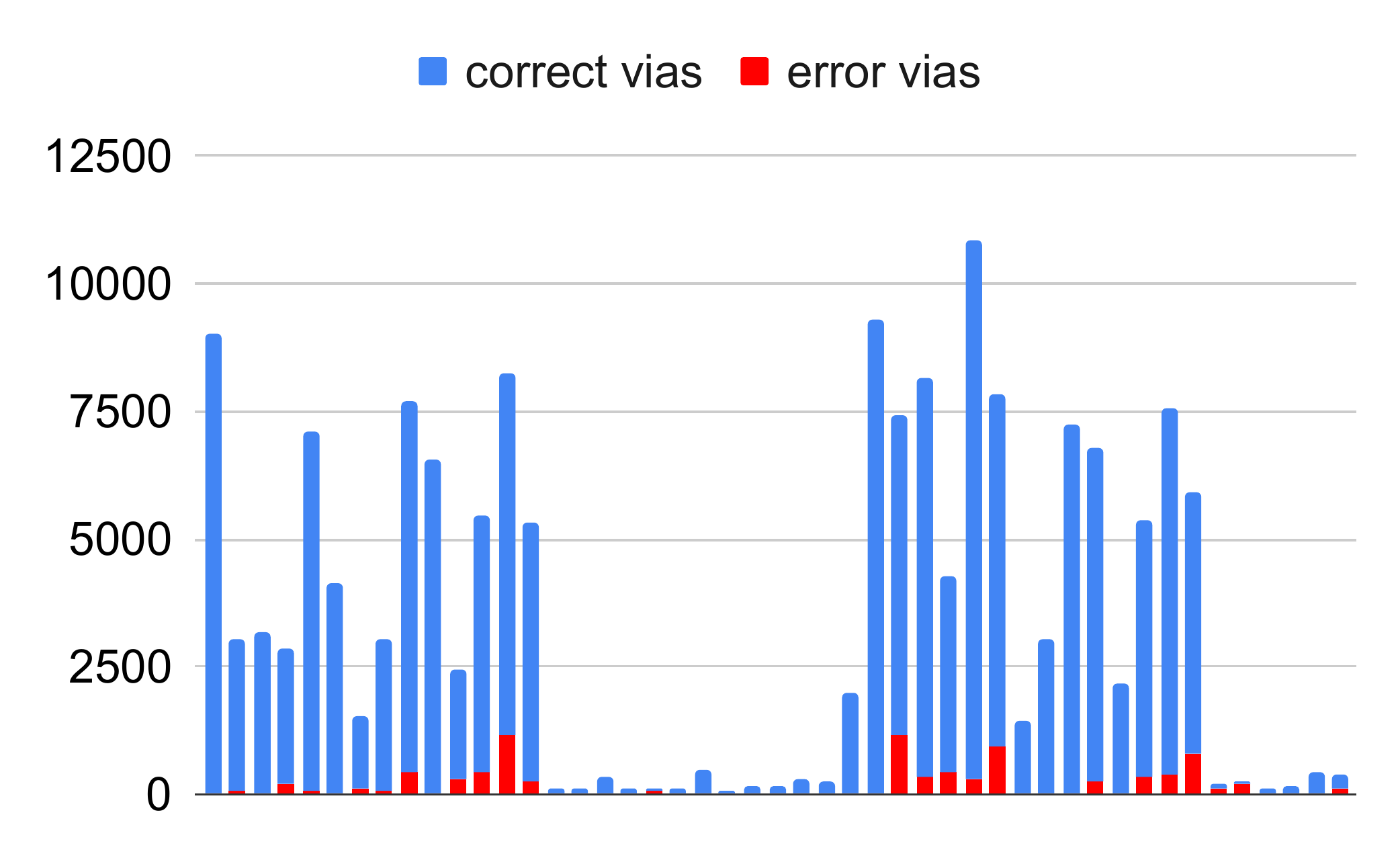}
    \caption{The statistics of correct vias and error vias in the testing set.}
    \label{viadataset}
\end{figure}
We alternately take V0 and V1 as the ground truth set and error set for testing. The via images with no errors are ruled out. The statistics of correct vias and error vias in the testing set are shown in Fig.~\ref{viadataset}. The number of correct vias and error vias vary dramatically for different images. The average ratio of via errors is 0.09. The experiment results of error detection are shown in Fig.~\ref{viaresult}. 

We achieve an average recall/precision of 0.96/0.90 on the testing set, which shows that our approach can effectively detect via errors. We further look into the only outliner result in Fig.~\ref{viaresult}, which have 956 wrong detections and significantly higher than the rest of the results. It turns out that most of the wrong detections on this image are false positives caused by the low contrast between vias and wires. Because of the low contrast, the estimate $v$ and $w$ values tend to be very close, which results in a failure of filtering. Although such samples are rare in our dataset, this result does reveal one of the limitations of our approach. We will leave the problem of tackling such cases to future work. 
\section{Conclusion}
In this work, we propose an automatic error detection approach for segmented SEM images of IC circuits. We first formulate wire and via error detection problems into an image classification and image translation problem, respectively. By adapting existing approaches in these two domains with necessary image pre-processing and post-processing techniques, we achieve an average recall/precision of 0.92/0.93 in wire error detection and 0.96/0.90 in via error detection, respectively. Our approach significantly unblocks one of the major bottlenecks for automatic SEM image segmentation approaches. The evaluation is conducted on a real industrial dataset, which implies the broad applicability of our approach to real-world applications.

\bibliographystyle{IEEEbib}
\bibliography{refs}
\end{document}